\documentclass[10pt, conference, compsocconf, letterpaper]{IEEEtran}
\IEEEoverridecommandlockouts
\usepackage{todonotes}

\usepackage{cite}
\usepackage{amsmath,amssymb,amsfonts}
\DeclareMathOperator*{\argmax}{arg\,max}
\usepackage[ruled,vlined,linesnumbered]{algorithm2e}
\usepackage{algpseudocode} 
\usepackage{graphicx}
\usepackage{textcomp}
\usepackage{subfig}
\usepackage{multirow}
 \def\BibTeX{{\rm B\kern-.05em{\sc i\kern-.025em b}\kern-.08em
    T\kern-.1667em\lower.7ex\hbox{E}\kern-.125emX}}
\usepackage{colortbl}
\usepackage{soul}
\usepackage{dsfont}
\usepackage{bbm}
\usepackage{graphicx} 
\usepackage[colorlinks=true, allcolors=blue]{hyperref}%

\begin{document}

\definecolor{mintgreen}{rgb}{0.6, 1.0, 0.6}
\definecolor{pastelviolet}{rgb}{0.8, 0.6, 0.79}
\definecolor{peridot}{rgb}{0.9, 0.89, 0.0}
\definecolor{richbrilliantlavender}{rgb}{0.95, 0.65, 1.0}
\definecolor{robineggblue}{rgb}{0.0, 0.8, 0.8}

\definecolor{green}{rgb}{0.1,0.1,0.1}
\newcommand{\done}{\cellcolor{teal}done}

\title{\text EnCoMP: Enhanced Covert Maneuver Planning with Adaptive Threat-Aware Visibility Estimation using Offline Reinforcement Learning}

\author{Jumman Hossain$^{1}$, Abu-Zaher Faridee$^{1}$, Nirmalya Roy$^{1}$, Jade Freeman$^{2}$, Timothy Gregory$^{2}$, and Theron Trout$^{3}$%
\thanks{*This work has been supported by U.S. Army Grant \texttt{\#W911NF2120076}}%
\thanks{$^{1}$Jumman Hossain, Abu-Zaher Faridee, and Nirmalya Roy are with the Department of Information Systems, University of Maryland, Baltimore County, USA. {\tt\small \{jumman.hossain, faridee1, nroy\}@umbc.edu}}%
\thanks{$^{2}$Jade Freeman and Timothy Gregory are with the DEVCOM Army Research Lab, USA. {\tt\small \{jade.l.freeman2.civ, timothy.c.gregory6.civ\}@army.mil}}%
\thanks{$^{3}$Theron Trout is with Stormfish Scientific Corp. {\tt\small theron.trout@stormfish.io}}%
}
% \author{\IEEEauthorblockN{Jumman Hossain$^1$, Nirmalya Roy$^1$}
% \IEEEauthorblockA{	
% $^1$Center for Real-time Distributed Sensing and Autonomy}
% \IEEEauthorblockA{$^1$Department of Information Systems,
% University of Maryland, Baltimore County, USA}
% % \IEEEauthorblockA{$^2$Army Research Lab, USA}
% \IEEEauthorblockA{$^1$\{jumman.hossain, nroy\}@umbc.edu}}

\maketitle
\begin{abstract}

Autonomous robots operating in complex environments face the critical challenge of identifying and utilizing environmental cover for covert navigation to minimize exposure to potential threats. We propose \textit{EnCoMP}, an enhanced navigation framework that integrates offline reinforcement learning and our novel Adaptive Threat-Aware Visibility Estimation (ATAVE) algorithm to enable robots to navigate covertly and efficiently in diverse outdoor settings. ATAVE is a dynamic probabilistic threat modeling technique that we designed to continuously assess and mitigate potential threats in real-time, enhancing the robot's ability to navigate covertly by adapting to evolving environmental and threat conditions. Moreover, our approach generates high-fidelity multi-map representations, including cover maps, potential threat maps, height maps, and goal maps from LiDAR point clouds, providing a comprehensive understanding of the environment. These multi-maps offer detailed environmental insights, helping in strategic navigation decisions. The goal map encodes the relative distance and direction to the target location, guiding the robot's navigation. We train a Conservative Q-Learning (CQL) model on a large-scale dataset collected from real-world environments, learning a robust policy that maximizes cover utilization, minimizes threat exposure, and maintains efficient navigation. We demonstrate our method’s capabilities on a physical Jackal robot, showing extensive experiments across diverse terrains. These experiments demonstrate \textit{EnCoMP}'s superior performance compared to state-of-the-art methods, achieving a 95\% success rate, 85\% cover utilization, and reducing threat exposure to 10.5\%, while significantly outperforming baselines in navigation efficiency and robustness.

\end{abstract}
\section{Introduction}
\label{sec:intro}

Autonomous navigation in complex environments is a critical capability for robots operating in various applications, such as military reconnaissance ~\cite{reconn1}, search and rescue missions ~\cite{sar1}, and surveillance operations ~\cite{surv1}. These scenarios pose unique challenges for robots, requiring them to accurately perceive the environment, identify potential cover, and adapt their navigation strategies to minimize exposure to threats while efficiently reaching the goal. In this context, a threat is any external factor capable of recognizing or interfering with the agent's presence and objectives. Developing robust navigation strategies that can effectively navigate in these environments while maintaining covertness is a challenging task, as it requires accounting for various uncertainties, dynamic obstacles, and environmental factors. Existing approaches to covert navigation often rely on pre-defined environmental models ~\cite{predef1, predef2} or supervised learning techniques ~\cite{sup1, sup2} that require extensive manual annotation and labeling of traversable terrain. However, these methods may not align with the actual traversability capabilities of different robots due to varying dynamic constraints, leading to conservative and inefficient navigation behaviors ~\cite{conserv1, conserv2}.

\begin{figure}[!htb]
    \centering
    \includegraphics[width=\linewidth]{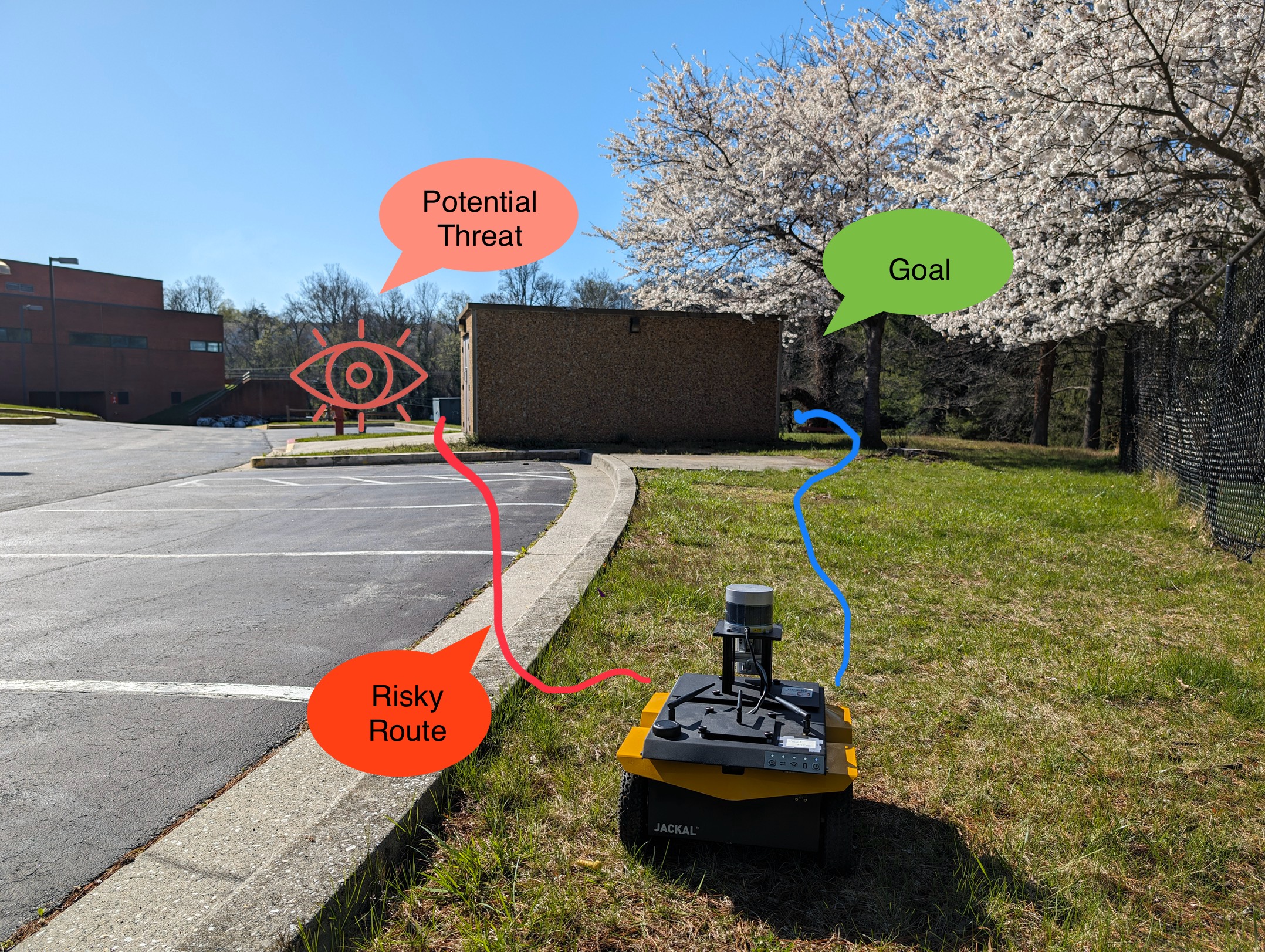}
    \caption{
    \textit{EnCoMP Covert Navigation Strategies}: 
In this outdoor environment, the robot (Jackal) is navigating towards a goal location. To achieve this while minimizing the risk of detection or interference from potential threats—defined as any external factors capable of recognizing or interfering with the robot's presence and objectives (represented by the eye icons)—the robot identifies and strategically moves toward nearby trees and a small brick building structure that can provide cover and concealment, instead of taking the risky route through the open area. By utilizing these natural and artificial features in the environment, the robot reduces its visibility and exposure as it traverses the field to reach its destination safely.}
   \label{fig}
\end{figure}

Reinforcement learning (RL) is a promising approach for autonomous navigation, enabling robots to learn complex strategies through interaction with the environment ~\cite{rl1, rl2}. Online RL methods have been applied to outdoor navigation ~\cite{online1, online2}, eliminating the need for human labeling. However, sim-to-real transfer issues ~\cite{sim2real1, sim2real2, weerakoon2022sim} often lead to severe performance degradation during real-world deployment of models trained using online reinforcement learning. In general, training complex models using online RL requires high-fidelity simulations, which may not be available for intricate environments. The discrepancies between the simulated and real-world domains can result in suboptimal performance when the trained models are applied to real-world scenarios. Offline RL ~\cite{offline1, offline2} has been proposed to mitigate these limitations by training models using data collected in real-world environments, reducing sim-to-real transfer problems. Despite the advancements in RL-based navigation, most existing methods have not fully exploited the rich information provided by modern sensors, such as LiDAR, which can offer valuable insights into the environment's properties and potential cover ~\cite{lidar1, lidar2}.

To address these challenges, we propose an enhanced covert navigation framework that leverages LiDAR data, height maps, cover maps, potential threat maps, and offline reinforcement learning. Our approach enables robots to identify and utilize natural and artificial environmental features as cover, thereby minimizing exposure to potential threats while maintaining efficient navigation.

The main contributions of our work are as follows:

\begin{itemize}
  \item \textbf{Novel Offline Reinforcement Learning Framework for Covert Navigation:}
  We propose a novel offline reinforcement learning-based framework, \textit{EnCoMP}, to learn a Q-function that evaluates a ground robot's candidate actions in terms of their ability to maximize cover utilization, minimize threat exposure, and reach the goal efficiently in complex environments. The framework incorporates a state-of-the-art Conservative Q-Learning (CQL) model ~\cite{kumar2020conservative} that leverages LiDAR-based perception to capture rich environmental information. Our model is trained using a diverse real-world dataset collected in various environments, which is automatically processed to generate states, actions, and rewards. This offline learning approach mitigates the sim-to-real transfer issues present in existing reinforcement learning methods for covert robot navigation. \textit{EnCoMP} demonstrates a significant improvement of up to 20\% in terms of navigation success rate compared to existing approaches.

  \item \textbf{Mapping for Strategic Navigation Decision-Making:}
   We design an observation space that captures rich environmental information to facilitate intelligent decision-making. We introduce a novel combination of robot-centric height maps, cover maps, and threat maps generated from high-resolution LiDAR point cloud data. These maps provide detailed representations of terrain elevation, cover object presence and quality, and potential threat positions and line-of-sight, enabling the robot to assess traversability, strategically utilize cover, and proactively avoid threats. Additionally, we incorporate a goal map that encodes the relative distance and direction to the target location, guiding efficient navigation. This novel integration of LiDAR-based observations empowers \textit{EnCoMP} to make informed decisions prioritizing covert operation and goal-reaching.

  \item \textbf{Adaptive Threat-Aware Visibility Estimation (ATAVE) Algorithm:}
We introduce a dynamic probabilistic threat modeling technique, Adaptive Threat-Aware Visibility Estimation (ATAVE) Algorithm. This algorithm adapts in real-time to changes in the robot's sensory observations, providing a continuously updated assessment of potential environmental threats. ATAVE incorporates advanced visibility computation methods, including efficient line-of-sight checks and a prioritized tree structure for rapid threat evaluation. This allows for enhanced situational awareness and significantly improves decision-making under uncertain and dynamic conditions.

 \item \textbf{Performance Improvement:}
  We evaluate \textit{EnCoMP} in three different real-world outdoor covert navigation scenarios, including urban, forested, and mixed terrains. We measure the success rate, cover utilization, and threat exposure. We also compare the results with our previous work, CoverNav, which serves as a baseline. \textit{EnCoMP} achieves a 20\% improvement in average success rate, 25\% improvement in average cover utilization, and 15\% reduction in average threat exposure across the three covert navigation scenarios.
\end{itemize}

\vspace{1mm}

The remainder of this paper is organized as follows: Section \ref{related_work} provides an overview of related work in outdoor navigation. Some related background information is present in Section \ref{background} Section \ref{encomp_approach} presents our proposed covert navigation framework, detailing the perception pipeline, offline reinforcement learning algorithm, and policy transfer to a physical robot platform. Section \ref{sec:experiments} describes the experimental setup and presents the results of our real-world tests. Finally, Section \ref{conclusion} concludes the paper and discusses potential future research directions.
\section{Related work}
\label{related_work}
In this section, we review the current research on perception in outdoor environments and explore offline RL methods applied to navigation tasks. Finally, we discuss the existing methods for covert navigation planning.
\subsection{Perception for Outdoor Navigation}
Accurate perception of the environment is crucial for autonomous navigation in complex outdoor scenes. Traditional approaches rely on supervised learning techniques, such as semantic segmentation \cite{sathyamoorthy2022terrapn} and object detection \cite{sathyamoorthy2020frozone}, to identify traversable terrain and potential cover. However, these methods require extensive manual annotation and may not generalize well to new environments or robot platforms \cite{papadakis2013terrain, weerakoon2022terp}.
Recent works have explored self-supervised learning \cite{sathyamoorthy2022terrapn} and adversarial training \cite{fan2020learning} to reduce the reliance on human annotation. While these techniques have shown promise, they still struggle to fully capture the complex properties and dynamics of real-world environments.

% In contrast to these approaches, our framework leverages multi-modal sensory information, specifically LiDAR intensity and height data, to generate high-fidelity cover maps and potential threat maps. This allows the robot to accurately perceive and reason about the environment's properties and potential cover without relying on extensive manual annotation.

In contrast to these approaches, our framework leverages the rich information provided by LiDAR point cloud data to generate high-fidelity height maps, cover maps, and threat maps. By processing the LiDAR data, our framework enables the robot to accurately perceive and reason about the environment's properties, including terrain elevation, potential cover locations, and threat positions, without relying on extensive manual annotation or simplified representations. 

% This LiDAR-based perception approach allows EnCoMP to capture detailed environmental information and facilitate intelligent decision-making for covert navigation in complex outdoor environments.

\subsection{Offline Reinforcement Learning for Navigation}
Offline reinforcement learning is a promising approach for learning navigation policies from previously collected datasets, mitigating the challenges associated with online learning in real-world environments \cite{levine2020offline, fujimoto2019off}. Recent works have demonstrated the effectiveness of offline reinforcement learning for navigation tasks in simulated environments \cite{mandlekar2020iris, hansen2021generalization}. Hansen et al. \cite{hansen2021generalization} introduced a generalization through offline reinforcement learning framework that learns a navigation policy from a diverse dataset of trajectories in simulated environments. Shah et al.  presented ReViND ~\cite{shah2023offline}, the first offline RL system for robotic navigation that can leverage previously collected data to optimize user-specified reward functions in the real-world. The system is evaluated for off-road navigation without any additional data collection or fine-tuning, and it is shown that it can navigate to distant goals using only offline training from this dataset. Another work by Li et al. ~\cite{hao2023avoidance} proposes an efficient offline training strategy to speed up the inefficient random explorations in the early stage of navigation learning.

Our approach builds upon these advancements by introducing an offline reinforcement learning framework specifically designed for covert navigation in complex outdoor environments. By learning from a diverse dataset collected from real-world settings, our approach can effectively capture the complex dynamics and uncertainties present in these environments and generate robust navigation strategies.

\subsection{Covert Navigation}
Covert navigation is a critical capability for robots operating in hostile or sensitive environments, where the primary objective is to reach a designated goal while minimizing the risk of detection by potential threats \cite{ajai2019survey, niroui2019deep}. Existing approaches to stealth navigation often rely on heuristic methods \cite{tews2004avoiding} or potential field-based techniques \cite{marzouqideceptive} to identify potential cover and minimize exposure. However, these methods may not effectively capture the complex dynamics and uncertainties present in real-world environments. Recent works have explored the use of reinforcement learning for stealth navigation ~\cite{mendoncca2018reinforcement}. While these approaches have shown promise in learning adaptive strategies, they often rely on simplified environmental representations and do not fully exploit the rich sensory information available in real-world settings.

Our framework addresses these limitations by integrating LiDAR-based perception and offline reinforcement learning, enabling robots to navigate covertly and efficiently in complex outdoor environments. By leveraging the rich environmental information captured by LiDAR sensors and training a robust policy using a diverse real-world dataset, our framework significantly improves navigation success rates, cover utilization, and threat avoidance compared to existing methods.
\section{Background}
\label{background}
In this section, we present our problem formulation and provide a concise overview of the offline reinforcement learning framework utilized for policy learning. We also detail the symbols and notations summarized in Table \ref{tab:symbols_definitions}.

\subsection{Problem Formulation}

We formulate the covert navigation problem as a Markov Decision Process (MDP) defined by the tuple $(\mathcal{S}, \mathcal{A}, P, R, \gamma)$, where $\mathcal{S}$ is the state space, $\mathcal{A}$ is the action space, $P$ is the transition probability function, $R$ is the reward function, and $\gamma$ is the discount factor. The state space $\mathcal{S}$ includes the robot's position $s_r \in \mathbb{R}^2$, the goal position $s_g \in \mathbb{R}^2$, the height map $H \in \mathbb{R}^{M \times N}$, the cover map $C \in \mathbb{R}^{M \times N}$, and the threat map $T \in \mathbb{R}^{M \times N}$, where $M$ and $N$ are the dimensions of the maps. The action space $\mathcal{A}$ consists of continuous robot navigation actions, such as linear and angular velocities. The transition probability function $P(s' | s, a)$ defines the probability of transitioning from state $s$ to state $s'$ when taking action $a$. The reward function $R(s, a)$ assigns a scalar value to each state-action pair, taking into account the robot's distance to the goal, the level of cover utilized, and the exposure to potential threats. The discount factor $\gamma \in [0, 1]$ determines the importance of future rewards compared to immediate rewards.
The objective is to learn an optimal policy $\pi^*: \mathcal{S} \rightarrow \mathcal{A}$ that maximizes the expected cumulative discounted reward:

\begin{equation}
\pi^* = \argmax_{\pi} \mathbb{E}^\pi \left[ \sum_{t=0}^{\infty} \gamma^t r_t \right]
\end{equation}

where $r_t$ is the reward received at time step $t$. \textit{EnCoMP} learns this optimal policy using offline reinforcement learning, specifically Conservative Q-Learning (CQL) ~\cite{kumar2020conservative}, which addresses the challenges of learning from a fixed dataset while ensuring conservative behavior when encountering novel or underrepresented state-action pairs during deployment.

\begin{table}[ht]
\centering

\begin{tabular}{|c|p{6cm}|}
\hline
\textbf{Symbol} & \textbf{Definition} \\ \hline
$\lambda_c$ & Weight for cover utilization reward \\ \hline
$\lambda_t$ & Weight for threat exposure penalty \\ \hline
$\lambda_g$ & Weight for goal-reaching reward \\ \hline
$\lambda_o$ & Weight for collision avoidance penalty \\ \hline
$r$ & Robot position \\ \hline
$c_i$ & Cell in the environment \\ \hline
$\mathcal{C}$ & Set of all cells in the environment \\ \hline
$\mathcal{T}_{path}$ & Planned trajectory \\ \hline
$\mathcal{M}_c$ & Cover map \\ \hline
$\mathcal{M}_g$ & Goal map \\ \hline
$p_t(c_i)$ & Probabilistic threat model for cell $c_i$ \newline at time $t$ \\ \hline
$v(c_i, r)$ & Visibility of cell $c_i$ from position $r$ \\ \hline
$\tau(c_i, r)$ & Multi-perspective threat assessment of \newline cell $c_i$ from position $r$ \\ \hline
$\rho(c_i, \mathcal{T}_{path})$ & Temporal visibility prediction of \newline cell $c_i$ along trajectory $\mathcal{T}_{path}$ \\ \hline
$\phi(c_i, \mathcal{T}_{path})$ & Cover-aware threat assessment of \newline cell $c_i$ along trajectory $\mathcal{T}_{path}$ \\ \hline
\end{tabular}
\caption{List of symbols used in our approach}
\label{tab:symbols_definitions}
\end{table}

\subsection{Offline Reinforcement Learning}

In the context of our covert navigation problem, we face the challenge of learning an effective policy from a fixed dataset without further interaction with the environment. While offline reinforcement learning offers a promising approach, directly applying standard reinforcement learning algorithms to offline datasets can lead to suboptimal performance. This is due to several factors, such as the mismatch between the distribution of states and actions in the dataset and the distribution induced by the learned policy, as well as the potential for overestimation of Q-values for unseen state-action pairs ~\cite{prudencio2023survey}.

To address these challenges, we employ the Conservative Q-Learning (CQL) algorithm ~\cite{kumar2020conservative} as our offline reinforcement learning method. CQL is an off-policy algorithm that aims to learn a conservative estimate of the Q-function by incorporating a regularization term in the training objective. The key idea behind CQL is to encourage the learned Q-function to assign lower values to unseen or out-of-distribution actions while still accurately fitting the Q-values for actions observed in the dataset. The CQL algorithm modifies the standard Q-learning objective by adding a regularization term that penalizes overestimation of Q-values. This regularization term is based on the difference between the expected Q-value under the learned policy and the expected Q-value under the behavior policy used to collect the dataset. By minimizing this difference, CQL encourages the learned Q-function to be conservative in its estimates, especially for state-action pairs that are not well-represented in the dataset.
\section{EnCoMP: Enhanced Covert Maneuver Planning}
\label{encomp_approach}
\begin{figure}[h]
    \centering
    % Syntax: \includegraphics[trim={left bottom right top},clip,width=\linewidth]{filename.pdf}
    \includegraphics[trim=0cm 0cm 2cm 2cm, clip,width=\linewidth]{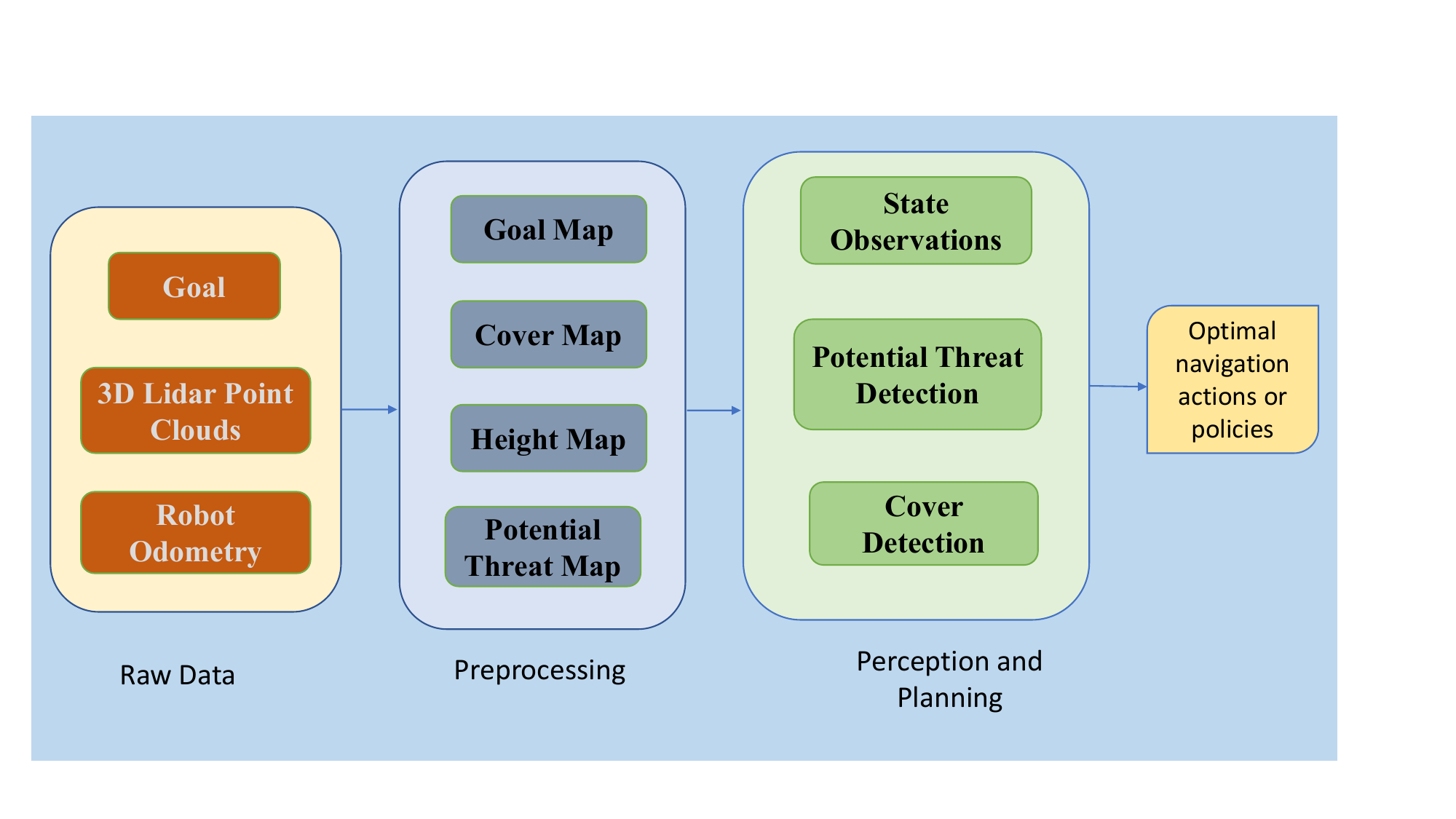}
    \caption{Overview of \textit{EnCoMP} System Architecture.} 
    \label{fig:architecture}
\end{figure}

In this section, We explain the major stages of the \textit{EnCoMP} approach. Fig. \ref{fig:architecture} shows how different modules in our method are connected. 

\subsection{Dataset Generation}
Our raw training data is collected by navigating a ground robot equipped with a 3D LiDAR sensor through various outdoor environments for approximately 6 hours. The robot is teleoperated to explore diverse terrains, including open fields, wooded areas, and urban settings with a mix of natural and artificial cover objects. During data collection, we record raw 3D point clouds, the robot's odometry, and GPS coordinates at each time step. Hence, the raw data set does not have any goal conditioning or goal-reaching policy, focusing solely on capturing varied environmental interactions without predefined navigation objectives.

To create a goal-conditioned dataset $\mathcal{D}$ with a series of $\{\mathbf{s}_j, \mathbf{a}_j, r_j, \mathbf{s}'_j\}$, we first process the raw point cloud data to generate height maps $\mathcal{H}$, cover maps $\mathcal{C}$, and threat maps $\mathcal{T}$ for each time step. The height map $\mathcal{H}$ represents the elevation of the terrain, the cover map $\mathcal{C}$ indicates the presence and quality of cover objects, and the threat map $\mathcal{T}$ represents the estimated positions and line-of-sight of potential threats in the environment. Next, we sample random trajectory segments from the processed dataset. For each segment, we select a random initial state $\mathbf{s}_j$ and a future state $\mathbf{s}'_j$ as the goal state, ensuring that the goal state is within a reasonable distance (i.e., 10-30 meters) from the initial state. The action $\mathbf{a}_j$ is derived from the robot's odometry and represents the motion command executed to transition from state $\mathbf{s}_j$ to the next state in the original trajectory. To introduce diversity in the goal locations and threat distributions, we augment the dataset by applying random rotations and translations to the height maps $\mathcal{H}$, cover maps $\mathcal{C}$, and threat maps $\mathcal{T}$. This augmentation helps to improve the generalization capability of the learning algorithm. The final processed dataset $\mathcal{D} = \{(\mathbf{s}_j, \mathbf{a}_j, r_j, \mathbf{s}'_j) \mid j = 1, 2, \ldots, N\}$ consists of $N$ goal-conditioned trajectories, where each trajectory is represented by a sequence of state-action-reward-next\_state tuples. The state $\mathbf{s}_j$ includes the height map $\mathcal{H}_j$, cover map $\mathcal{C}_j$, threat map $\mathcal{T}_j$, and the robot's position $\mathbf{p}_j$ at time step $j$. The next state $\mathbf{s}'_j$ represents the state reached after executing action $\mathbf{a}_j$ in state $\mathbf{s}_j$. This processed dataset is then used to train the covert navigation policy using offline reinforcement learning. The policy learns to generate actions that maximize the expected cumulative reward, taking into account the terrain height, cover availability, and threat exposure.

% By incorporating height maps, cover maps, and threat maps into the dataset generation process, we provide the learning algorithm with a rich representation of the environment, enabling it to learn effective strategies for covert navigation in complex outdoor scenes.

\subsection{Finding Cover in the Environment}
To identify cover objects in the environment, we process the LiDAR point cloud $\mathcal{P} = \{\mathbf{p}_i \in \mathbb{R}^3 \mid i = 1, \dots, N\}$, where each point $\mathbf{p}_i$ is represented by its 3D coordinates $(x_i, y_i, z_i)$. We apply the following steps:

\textbf{Point Cloud Segmentation:} We use the Euclidean cluster extraction algorithm to segment the point cloud into distinct objects. The algorithm groups nearby points based on their Euclidean distance, forming clusters $\mathcal{C}_j \subset \mathcal{P}$, where $j = 1, \dots, M$ and $M$ is the number of clusters.

\textbf{Cover Object Identification:} For each cluster $\mathcal{C}_j$, we calculate its height $h_j$, density $d_j$, and volume $v_j$ as follows:
\begin{align}
h_j &= \max_{\mathbf{p}_i \in \mathcal{C}_j} z_i - \min_{\mathbf{p}_i \in \mathcal{C}_j} z_i, \\
d_j &= \frac{|\mathcal{C}_j|}{v_j}, \\
v_j &= (x_{j,\max} - x_{j,\min}) \times (y_{j,\max} - y_{j,\min}) \times h_j,
\end{align}
where $|\mathcal{C}_j|$ is the number of points in cluster $\mathcal{C}_j$, and $x_{j,\min}, x_{j,\max}, y_{j,\min}, y_{j,\max}$ are the minimum and maximum coordinates of the cluster in the $x$ and $y$ dimensions, respectively. A cluster $\mathcal{C}_j$ is considered a cover object if it satisfies the following criteria:
\begin{align}
h_j &\geq h_{\min}, \\
d_j &\geq d_{\min}, \\
v_j &\geq v_{\min},
\end{align}
where $h_{\min}, d_{\min}, v_{\min}$ are predefined thresholds for height, density, and volume, respectively where $|\mathcal{C}_j|$ is the number of points in cluster $\mathcal{C}_j$, and $x_{j,\min}, x_{j,\max}, y_{j,\min}, y_{j,\max}$ are the minimum and maximum coordinates of the cluster in the $x$ and $y$ dimensions, respectively.

\subsection{Generating Cover Maps}
The cover map $\mathcal{M}_c$ is a 2D grid representation of the environment, where each cell $(i, j)$ contains the cover density value $c_{i,j} \in [0, 1]$. We generate the cover map using the identified cover objects as follows:

\begin{enumerate}
    \item Initialize the cover map $\mathcal{M}_c$ with dimensions $H \times W$ and cell size $\delta$.
    \item For each cover object $\mathcal{C}_j$, project its points onto the corresponding cells in the cover map. The projection of a point $\mathbf{p}_i = (x_i, y_i, z_i)$ to a cell $(i, j)$ is given by:
    \begin{align}
    i &= \left\lfloor \frac{x_i - x_{\min}}{\delta} \right\rfloor, \\
    j &= \left\lfloor \frac{y_i - y_{\min}}{\delta} \right\rfloor,
    \end{align}
    where $x_{\min}, y_{\min}$ are the minimum coordinates of the environment in the $x$ and $y$ dimensions, respectively.
    \item For each cell $(i, j)$ in the cover map, calculate the cover density $c_{i,j}$ as:
    \begin{equation}
    c_{i,j} = \frac{|\mathcal{P}_{i,j}^c|}{|\mathcal{P}_{i,j}|},
    \end{equation}
    where $\mathcal{P}_{i,j}$ is the set of all LiDAR points projected onto cell $(i, j)$, and $\mathcal{P}_{i,j}^c \subseteq \mathcal{P}_{i,j}$ is the subset of points belonging to cover objects.
\end{enumerate}

\subsection{Generating Height Maps}
The height map $\mathcal{M}_h$ is a 2D grid representation of the environment, where each cell $(i, j)$ contains the maximum height value $h_{i,j} \in \mathbb{R}$. We generate the height map using the LiDAR point cloud as follows:

\begin{enumerate}
    \item Initialize the height map $\mathcal{M}_h$ with the same dimensions and cell size as the cover map.
    \item For each LiDAR point $\mathbf{p}_i = (x_i, y_i, z_i)$, project it onto the corresponding cell $(i, j)$ in the height map using the projection equations defined earlier.
    \item For each cell $(i, j)$ in the height map, calculate the maximum height value $h_{i,j}$ as:
    \begin{equation}
    h_{i,j} = \max_{\mathbf{p}_i \in \mathcal{P}_{i,j}} z_i,
    \end{equation}
    where $\mathcal{P}_{i,j}$ is the set of all LiDAR points projected onto cell $(i, j)$.
\end{enumerate}

\subsection{Generating Goal Maps}
The goal map $\mathcal{M}_g$ is a 2D grid representation of the environment, where each cell $(i, j)$ contains a value indicating the relative distance and direction to the goal position. We generate the goal map as follows:
\begin{enumerate}
    \item Initialize the goal map $\mathcal{M}_g$ with the same dimensions and cell size as the cover map and height map.
    \item Set the goal position $(x_g, y_g)$ in the goal map.
    \item For each cell $(i, j)$ in the goal map, calculate the distance $d_{i,j}$ and angle $\theta_{i,j}$ to the goal position:
    \begin{equation}
    d_{i,j} = \sqrt{(x_i - x_g)^2 + (y_j - y_g)^2}
    \end{equation}
    \begin{equation}
    \theta_{i,j} = \text{atan2}(y_j - y_g, x_i - x_g)
    \end{equation}
    \item Normalize the distance values to the range $[0, 1]$ based on the maximum distance in the goal map.
    \item Encode the distance and angle information into the goal map $\mathcal{M}_g$, where each cell contains a tuple $(d_{i,j}, \theta_{i,j})$.
\end{enumerate}
The goal map provides the robot with information about the relative distance and direction to the goal position, guiding its navigation towards the target location.

\subsection{Adaptive Threat-Aware Visibility Estimation (ATAVE)}
\label{subsec:atave}
The Adaptive Threat-Aware Visibility Estimation (ATAVE) algorithm (see Algorithm~\ref{alg:atave}) is a pivotal component of the \textit{EnCoMP} system, designed to optimize navigation strategies by dynamically assessing and mitigating potential threats in real-time.

\begin{algorithm}
\caption{Adaptive Threat-Aware Visibility Estimation (ATAVE)}
\label{alg:atave}
\SetAlgoLined
\KwData{Environment $\mathcal{E}$, cells $\mathcal{C}$, robot position $r$, trajectory $\mathcal{T}_{\text{path}}$, cover map $\mathcal{M}_c$, goal map $\mathcal{M}_g$}
\KwResult{Threat assessment $\phi(c_i, \mathcal{T}_{\text{path}})$ for each $c_i$}
\textbf{Initialize:} $p_t(c_i)$ for all $c_i$\;
\textbf{Build Tree:} $\mathcal{T}_{\text{tree}}$ from threat probabilities\;
\ForEach{$c_i \in \mathcal{C}$}{
    Compute visibility $v(c_i, r)$ using LLA: $v(c_i, r) = \text{LLA}(\mathcal{T}_{\text{tree}}, c_i, r)$\;
    Assess threat $\tau(c_i, r)$: $\tau(c_i, r) = \max_{v_j} v(c_i, v_j) \cdot p_t(v_j)$\;
    Predict visibility $\rho(c_i, \mathcal{T}_{\text{path}})$: $\rho(c_i, \mathcal{T}_{\text{path}}) = \max_{r_k} \tau(c_i, r_k) \cdot \gamma^k$\;
    Evaluate threat $\phi(c_i, \mathcal{T}_{\text{path}})$: $\phi(c_i, \mathcal{T}_{\text{path}}) = \rho(c_i, \mathcal{T}_{\text{path}}) \cdot (1 - \mathcal{M}_c(c_i)) \cdot \mathcal{M}_g(c_i)$\;
}
\While{robot is navigating}{
    Update $p_t(c_i)$ with new data $o_t$: $p_t(c_i) = \frac{p_{t-1}(c_i) \cdot p(o_t | c_i)}{\sum_{c_j \in \mathcal{C}} p_{t-1}(c_j) \cdot p(o_t | c_j)}$\;
    Recompute $\phi(c_i, \mathcal{T}_{\text{path}})$ for all $c_i$\;
    Select next action based on $\phi(c_i, \mathcal{T}_{\text{path}})$\;
}
\end{algorithm}

\begin{figure}[!htb]
    \centering
    \includegraphics[width=\linewidth]{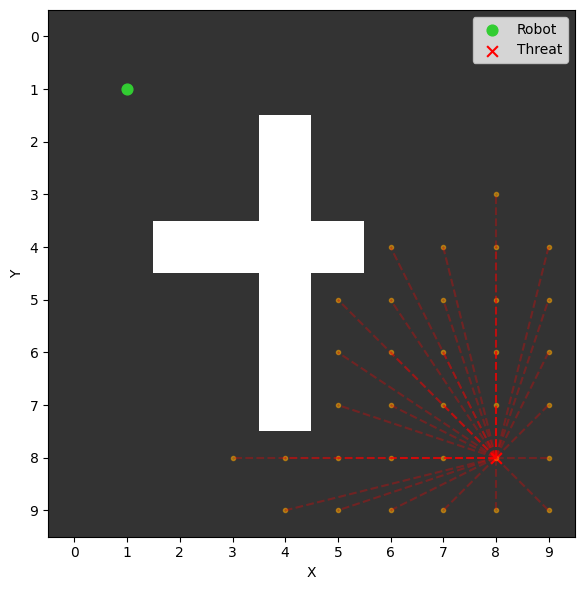}
    \caption{Visualization of the LLA* Visibility Calculation in the \textit{EnCoMP} framework. The plot shows the grid environment with obstacles (gray cells), the robot's position (green circle), the threat's position (red cross), and the visible cells (orange dots) determined by the Late Line-of-Sight Check and Prioritized Trees (LLA*) algorithm. The red dashed lines represent the line-of-sight from the threat to the visible cells within its visibility range. }
   \label{fig:llastar_visibility}
\end{figure}

At the core of ATAVE is a probabilistic threat model $p_t(c_i)$ maintained for each cell $c_i$ in the discretized environment $\mathcal{C}$. This model represents the likelihood of a threat being present in each cell at time $t$. The threat probabilities are initially estimated based on prior knowledge and are continuously updated using the robot's observations and sensor data through a Bayesian update process:
\begin{equation}
p_t(c_i) = \frac{p_{t-1}(c_i) \cdot p(o_t | c_i)}{\sum_{j=1}^{N} p_{t-1}(c_j) \cdot p(o_t | c_j)}
\end{equation}

where $p_{t-1}(c_i)$ is the prior probability of a threat in cell $c_i$ at time $t-1$, $p(o_t | c_i)$ is the likelihood of observing data $o_t$ given a threat in cell $c_i$, and the denominator ensures normalization.
To provide a comprehensive assessment of the risk associated with each cell, ATAVE employs a multi-perspective threat estimation approach. It considers the robot's visibility from multiple vantage points, denoted as $\mathcal{V} = \{v_1, v_2, \ldots, v_M\}$, representing potential threat perspectives. The multi-perspective threat assessment is computed as:
\begin{equation}
\tau(c_i, r) = \max_{v_j \in \mathcal{V}} v(c_i, v_j) \cdot p_t(v_j)
\end{equation}

where $\tau(c_i, r)$ represents the multi-perspective threat assessment of cell $c_i$ from the robot's position $r$, $v(c_i, v_j)$ is the visibility of cell $c_i$ from vantage point $v_j$, and $p_t(v_j)$ is the probability of a threat being present at vantage point $v_j$.
To enable real-time threat assessment, ATAVE utilizes efficient visibility computation techniques, such as the Late Line-of-Sight Check and Prioritized Tree Structures (LLA) \cite{Zhang2019LateLC}. These techniques significantly reduce the computational overhead while maintaining high accuracy in determining the robot's visibility from different threat perspectives. As shown in Figure~\ref{fig:llastar_visibility}, the LLA* algorithm efficiently computes the visibility of cells from the threat's perspective, enabling the robot to assess the risk of detection and make informed decisions during navigation.

The goal map $\mathcal{M}_g$ is also utilized in the ATAVE algorithm to guide the robot's navigation while considering the threat exposure. The cover-aware threat assessment for a cell $c_i$ along the robot's planned trajectory $\mathcal{T}$ is computed as:
\begin{equation}
\phi(c_i, \mathcal{T}) = \rho(c_i, \mathcal{T}) \cdot (1 - \mathcal{M}_c(c_i)) \cdot \mathcal{M}_g(c_i)
\end{equation}

where $\phi(c_i, \mathcal{T})$ represents the cover-aware threat assessment of cell $c_i$ along trajectory $\mathcal{T}$, $\rho(c_i, \mathcal{T})$ is the temporal visibility prediction of cell $c_i$ along trajectory $\mathcal{T}$, $\mathcal{M}_c(c_i)$ is the cover density of cell $c_i$, and $\mathcal{M}_g(c_i)$ is the goal map value of cell $c_i$, indicating its proximity and direction to the goal.
By incorporating the goal map into the cover-aware threat assessment, ATAVE prioritizes navigation decisions that balance the objectives of minimizing threat exposure, maximizing cover utilization, and progressing towards the goal. 

% This integration allows the robot to make informed decisions that adapt to the complex environment while maintaining a goal-oriented behavior.
% Based on the dynamic threat assessment, ATAVE continuously adapts the robot's navigation strategy to minimize exposure to high-risk areas. It integrates with the cover utilization and path planning components of \textit{EnCoMP} to generate safer and more efficient trajectories that prioritize covert operations.

\textbf{Proposition V.1.} \textit{The ATAVE algorithm modifies the threat exposure function \(\varphi(c_i, T_{path})\) such that traversability costs in high-threat regions are always higher than those in low-threat regions.} 

\textbf{Proof.} Consider the cover-aware threat assessment function \(\varphi(c_i, T_{path})\) used in the ATAVE algorithm: \(\varphi(c_i, T_{path}) = \rho(c_i, T_{path}) \cdot (1 - M_c(c_i)) \cdot M_g(c_i)\) where \(\rho(c_i, T_{path})\) is the temporal visibility prediction of cell \(c_i\) along the trajectory \(T_{path}\), \(M_c(c_i)\) is the cover density of cell \(c_i\), and \(M_g(c_i)\) is the goal map value of cell \(c_i\). The maximum cost for a region with low cover density (high visibility and threat probability) can be expressed as: \(\varphi_{\max} = \rho_{\max} \cdot (1 - M_{c_{\min}}) \cdot M_{g_{\max}}\) where \(\rho_{\max}\) is the maximum visibility prediction, \(M_{c_{\min}}\) is the minimum cover density (which is 0 for no cover), and \(M_{g_{\max}}\) is the maximum goal map value. The minimum cost for a region with high cover density (low visibility and threat probability) can be expressed as: \(\varphi_{\min} = \rho_{\min} \cdot (1 - M_{c_{\max}}) \cdot M_{g_{\min}}\) where \(M_{c_{\max}}\) is the maximum cover density (which is 1 for full cover), \(\rho_{\min}\) is the minimum visibility prediction, and \(M_{g_{\min}}\) is the minimum goal map value. Given \(M_{c_{\max}} = 1\) and assuming \(\rho_{\min}\) and \(M_{g_{\min}}\) are non-zero, this simplifies to: \(\varphi_{\min} = 0\). Since \(\rho_{\max} \cdot (1 - M_{c_{\min}}) \cdot M_{g_{\max}} > 0\), it follows that: \(\varphi_{\max} > \varphi_{\min}\). 

Therefore, the ATAVE algorithm assigns higher costs to regions with higher visibility and threat probability. Consequently, the agent, following the path of least cost, will prefer navigating through regions with higher cover density and lower visibility and threat probability. Thus, regions with lower threat exposure will always be preferred for navigation. \(\blacksquare\)

\subsection{Reward Functions}

We design the following reward functions to capture the objectives of the covert navigation problem:

\textbf{Cover Utilization Reward : }The cover utilization reward encourages the robot to navigate through areas with high cover density, as indicated by the cover map $\mathcal{C}$. It is defined as:

\begin{equation}
R_{\text{cover}}(s, a) = \lambda_c \sum_{i=1}^{M} \sum_{j=1}^{N} \mathcal{C}_{ij} \cdot \mathbbm{1}\left[(x_i, y_j) \in \xi(s, a)\right]
\end{equation}

where $\lambda_c$ is a positive weighting factor, $\mathcal{C}_{ij}$ is the cover density value at cell $(i, j)$ in the cover map, $\mathbbm{1}[\cdot]$ is the indicator function, and $\xi(s, a)$ is the set of cells traversed by the robot when taking action $a$ in state $s$.

\textbf{Threat Exposure Penalty : } The threat exposure penalty discourages the robot from being exposed to potential threats, as indicated by the potential threat map $\mathcal{T}$. It is defined as:

\begin{equation}
R_{threat}(s) = -\lambda_t \sum_{i=1}^{M} \sum_{j=1}^{N} \mathcal{T}_{ij} \cdot \mathbbm{1}[(x_i, y_j) \in \xi(s)]
\end{equation}

where $\lambda_t$ is a weighting factor, $\mathcal{T}_{ij}$ is the threat level at cell $(i, j)$ in the potential threat map, and $\xi(s)$ is the set of cells occupied by the robot in state $s$.

\textbf{Goal Reaching Reward : }The goal reaching reward encourages the robot to make progress towards the goal position. It is defined as:

\begin{equation}
R_{goal}(s, a) = \lambda_g (d(s, g) - d(s', g))
\end{equation}

where $\lambda_g$ is a positive weighting factor, $d(s, g)$ is the distance between the robot's position in state $s$ and the goal position $g$, and $s'$ is the state reached after taking action $a$ in state $s$.

\textbf{Collision Penalty : }The collision penalty discourages the robot from colliding with obstacles or navigating through untraversable regions, as indicated by the height map $\mathcal{H}$. It is defined as:

\begin{equation}
R_{\text{collision}}(s, a) = 
\begin{cases}
-\lambda_o, & \text{if } \exists (x_i, y_j) \in \xi(s, a), \mathcal{H}_{ij} > h_{\max}, \\
0, & \text{otherwise.}
\end{cases}
\end{equation}

where $\lambda_o$ represents the weight assigned to the collision penalty, $\mathcal{H}_{ij}$ is the height value at cell $(i, j)$ in the height map, and $h_{max}$ is the maximum traversable height threshold.

\subsection{CQL Policy Learning with Multi-Map Inputs}

We employ Conservative Q-Learning (CQL) ~\cite{kumar2020conservative} to learn the covert navigation policy from the diverse dataset $\mathcal{D}$ collected from real-world environments. CQL learns a conservative estimate of the Q-function by incorporating a regularization term in the training objective, which encourages the learned Q-function to assign lower values to out-of-distribution actions while still accurately fitting the Q-values for actions observed in the dataset. The CQL objective is defined as:

\begin{equation}
\begin{split}
\mathcal{L}(\theta) = \mathbb{E}_{(\mathbf{s}, \mathbf{a}, r, \mathbf{s}') \sim \mathcal{D}} \Big[ &(Q_\theta(\mathbf{s}, \mathbf{a}) - \\
&(r + \gamma \mathbb{E}_{\mathbf{a}' \sim \pi_\theta(\cdot \mid \mathbf{s}')}[Q_\theta(\mathbf{s}', \mathbf{a}')]))^2 \Big] \\
&+ \alpha \mathcal{R}(\theta)
\end{split}
\end{equation}

where $Q_\theta$ is the Q-function parameterized by $\theta$, $\pi_\theta$ is the policy, $\alpha$ is a hyperparameter controlling the strength of the regularization term $\mathcal{R}(\theta)$, and $\mathcal{D}$ is the dataset of transitions.

The regularization term $\mathcal{R}(\theta)$ is defined as:

\begin{equation}
\begin{split}
\mathcal{R}(\theta) = \mathbb{E}_{\mathbf{s} \sim \mathcal{D}} \bigg[ &\log \sum_{\mathbf{a}} \exp(Q_\theta(\mathbf{s}, \mathbf{a})) \\
&- \mathbb{E}_{\mathbf{a} \sim \pi_\beta(\cdot \mid \mathbf{s})}[Q_\theta(\mathbf{s}, \mathbf{a})] \bigg]
\end{split}
\end{equation}

where $\pi_\beta$ is the behavior policy used to collect the dataset $\mathcal{D}$. To learn the covert navigation policy, we use the high-fidelity cover map $\mathcal{C}$, potential threat map $\mathcal{T}$, height map $\mathcal{H}$, and goal map $\mathcal{M}_g$ as input to the Q-function and policy networks. The state input to the networks is defined as:
\begin{equation}
s = [\mathcal{C}, \mathcal{T}, \mathcal{H}, \mathcal{M}_g, p_r, p_g]
\end{equation}
where $p_r$ and $p_g$ are the robot's position and goal position, respectively.

During training, the CQL algorithm samples batches of transitions $(\mathbf{s}, \mathbf{a}, r, \mathbf{s}')$ from the dataset $\mathcal{D}$ and updates the Q-function and policy networks using the CQL objective. The Q-function network learns to estimate the expected cumulative reward for each state-action pair, while the policy network learns to select actions that maximize the Q-values. By incorporating the multi-map inputs and utilizing the conservative Q-learning objective, the learned policy effectively navigates the robot towards the goal while maximizing cover utilization, minimizing threat exposure, and avoiding collisions. The CQL algorithm ensures that the learned policy is robust and generalizes well to novel environments by penalizing overestimation of Q-values for out-of-distribution actions.

During deployment, the learned policy takes the current state $\mathbf{s}$, which includes the multi-map inputs, and outputs the optimal action $\mathbf{a}^*$ to navigate the robot covertly and efficiently. The robot executes the selected action and observes the next state $\mathbf{s}'$ and reward $r$. This process is repeated until the robot reaches the goal or a maximum number of steps is reached,  which serves as the failure condition. By leveraging the map inputs and the CQL algorithm, our approach learns a robust and efficient policy for covert navigation in complex environments, enabling the robot to make informed decisions based on the comprehensive understanding of the environment provided by the cover map, potential threat map, and height map.
% \begin{equation}
% \mathbf{a}^* = \pi_\theta(\mathbf{s})
% \end{equation}

\subsection{Threat-Aware Cover-Based Planning}

To effectively handle covert navigation in complex environments, we design a dynamic action selection strategy that utilizes an enhanced covert maneuver planner. For our robot, an action \(a\) from the set of all possible actions \(A\) is evaluated using \(Q_{\min}(s, a; \theta)\), where \(s\) is the current state and \(\theta\) represents the parameters of our model. The action space \(V_s\) is defined based on the robot's operational capabilities and the environmental context:
\[
V_s =
\begin{cases}
(v_x, \omega_z), & \text{if minimal threat is detected} \\
(v_x', \omega_z'), & \text{otherwise}
\end{cases}
\]
where \(v_x, v_x' \in [0, 2.0]\) m/s and \(\omega_z, \omega_z' \in [-1.5, 1.5]\) rad/s, reflecting the robot's maximum linear and angular velocities.

When minimal threats are detected, the robot can move at its normal speed and angular velocity range. However, in higher threat scenarios, the linear and angular velocities are reduced to \(v_x'\) and \(\omega_z'\), respectively, to minimize the robot's detectability and noise. The dynamic adjustment of velocities is based on the detected threat levels and the surrounding environment. In areas with higher threat levels or dense obstacles, the robot reduces its speed and angular velocity to maintain stealth and avoid detection. The  values of \(v_x'\) and \(\omega_z'\) are determined by functions that take into account the threat level and obstacle density:
\[
v_x' = f_v(\text{threat\_level}, \text{obstacle\_density})
\]
\[
\omega_z' = f_{\omega}(\text{threat\_level}, \text{obstacle\_density})
\]
where \(f_v\) and \(f_{\omega}\) are functions that map the \(\text{threat\_level} \in [0, 1]\)  and \(\text{obstacle\_density} \in [0, 1]\) to appropriate linear and angular velocity values, respectively. These functions are designed to ensure that the robot moves slowly and cautiously in high-threat and dense environments while still making progress towards the goal. Moreover, the system prioritizes paths that utilize natural landscape features to enhance covertness, adapting its movement patterns to the context of each specific mission scenario. The decision-making process involves selecting the action \(a^*\) that minimizes exposure to threats while maximizing progress towards the goal, calculated as:
\[
a^* = \arg\min_{a \in A} Q_{\min}(s, a).
\]
\section{Experiments and Results}
\label{sec:experiments}
In this section, we present the experimental setup, results, and analysis of the \textit{EnCoMP} framework. We evaluate the performance of \textit{EnCoMP} in real-world outdoor test scenarios that are not part of the original training dataset and compare it with state-of-the-art baselines.

\subsection{Evaluation Metrics}
We evaluate the performance of \textit{EnCoMP} and the baselines using the following metrics:

\begin{itemize}
\item \textbf{Success Rate}: The percentage of trials in which the robot successfully reaches the goal location without being detected by threats, colliding with obstacles, or running out of time.
\item \textbf{Navigation Time}: The average time taken by the robot to navigate from the start position to the goal location in successful trials.
\item \textbf{Trajectory Length}: The average length of the path traversed by the robot from the start position to the goal location in successful trials.

\item \textbf{Threat Exposure}: The average percentage of time the robot is exposed to threats during navigation, calculated as the ratio of the time spent in the line-of-sight of threats to the total navigation time.
\item \textbf{Cover Utilization}: The average percentage of time the robot utilizes cover objects during navigation, calculated as the ratio of the time spent in high cover density areas (as indicated by the cover map) to the total navigation time.
\end{itemize}

\subsection{Baselines}
We compare the performance of \textit{EnCoMP} against the following state-of-the-art navigation methods:
\begin{itemize}

   \item \textbf{CoverNav:} A deep reinforcement learning-based system designed for navigation planning that prioritizes the use of natural cover in unstructured outdoor environments. This strategy aims to enhance stealth and safety in potentially hostile settings ~\cite{hossain2023covernav}.
    
    \item \textbf{VAPOR:} A method for autonomous legged robot navigation in unstructured, densely vegetated outdoor environments using offline reinforcement learning ~\cite{weerakoon2023vapor}.
    
    \item \textbf{VERN:} An approach for vegetation-aware robot navigation, which facilitates efficient traversal in densely vegetated and unstructured outdoor environments. Their method integrates sensory data and learning algorithms to navigate with an understanding of vegetation density and type ~\cite{sathyamoorthy2023vern}.
\end{itemize}

\subsection{Implementation Details}
Our offline RL network is implemented in PyTorch and trained in a workstation with a 10th-generation Intel Core i9-10850K processor and an NVIDIA GeForce RTX 3090 GPU. For real-time deployment and inference, we use the Jackal UGV from Clearpath Robotics which runs on the Robot Operating System (ROS) Noetic distribution. The Jackal UGV is equipped with a 3D VLP-32C Velodyne Ultrapuck LiDAR,  an AXIS Fixed IP camera, featuring a fixed-focus lens with 30 fps/VGA resolution. The system’s processing capabilities are powered by an onboard Intel computer system, equipped with an Intel i7-9700TE CPU and an NVIDIA GeForce GTX 1650 Ti GPU.

The CQL algorithm is configured with a learning rate of \(1 \times 10^{-4}\), a batch size of 256, and a discount factor \(\gamma = 0.95\). The regularization term weight \(\alpha\) is set to 0.2 to balance exploration and exploitation. The neural networks used in the CQL algorithm employ ReLU activation functions for the hidden layers and a linear activation for the output layer. These hyperparameters were fine-tuned through extensive experimentation to optimize learning efficiency and ensure robust performance in covert navigation tasks.
The training process was conducted over 500 episodes to ensure stable convergence and robust performance of the learned policy.

To ensure safe navigation and obstacle avoidance, we also integrate the Dynamic Window Approach (DWA) ~\cite{fox1997dynamic} into the \textit{EnCoMP} framework. DWA is a local planner that generates velocity commands for the robot based on the current sensor readings and the robot's dynamics. It optimizes the velocity commands by considering the robot's current velocity, the obstacles in the environment, and the goal location.

% \subsection{Experimental Parameters and Settings}
% We provide detailed information on the experimental parameters and settings used in our evaluation:

% \begin{itemize}
% \item \textbf{Reward Weights:} The reward function weights were carefully tuned to balance the various objectives of the covert navigation task. Specifically, $\lambda_c = 0.5$ for the cover utilization reward, $\lambda_t = 1.5$ for the threat exposure penalty, $\lambda_g = 1.0$ for the goal-reaching reward, and $\lambda_o = 5.0$ for the collision avoidance penalty. These weight values were determined through extensive empirical evaluation to ensure effective and safe navigation while prioritizing threat avoidance and goal-reaching efficiency over cover utilization.
% \item \textbf{CQL Hyperparameters:} The Conservative Q-Learning (CQL) algorithm was configured with a learning rate of $1 \times 10^{-4}$, a batch size of 256, and a discount factor $\gamma = 0.95$. To achieve a balance between exploration and exploitation, the regularization term weight $\alpha$ was set to 0.2. The neural networks employed ReLU activation functions for the hidden layers and a linear activation for the output layer. These hyperparameter values were subsequently fine-tuned through extensive experimentation to optimize learning efficiency and ensure robust performance in our covert navigation tasks. The training process was conducted over 1,000 episodes to ensure stable convergence and robust performance of the learned policy
% \end{itemize}

\subsection{Testing Scenarios}
\label{sec:testingscenario}
We evaluate our covert-based navigation framework in three diverse outdoor environments. The three testing scenarios present increasing levels of difficulty for covert navigation:
% We evaluate our covert-based navigation framework in three diverse outdoor environments (see Fig. \ref{fig:environments_maps}):

\begin{itemize}
\item \textbf{Scenario 1}: An urban environment with buildings, structures, and sparse vegetation.
\item \textbf{Scenario 2}: A densely vegetated forest environment with trees, bushes, and uneven terrain.
\item \textbf{Scenario 3}: A mixed environment with both urban structures and natural vegetation.
\end{itemize}

Each scenario presents unique challenges for covert navigation, such as varying levels of cover availability, threat exposure, and terrain complexity.
\subsection{Performance Comparison and Analysis}
Table \ref{tab:real_results} demonstrates \textit{EnCoMP}'s significant improvements over existing navigation systems across all evaluation metrics and testing scenarios. Each scenario was tested with at least 10 runs to ensure the consistency and reliability of the results. In terms of success rate, \textit{EnCoMP} consistently outperforms the baseline methods across all scenarios, achieving success rates of 95\%, 93\%, and 91\% in Scenarios 1, 2, and 3, respectively. This demonstrates the effectiveness of our approach in navigating complex environments while maintaining covertness and avoiding threats. \textit{EnCoMP} also exhibits the shortest navigation times and trajectory lengths compared to the baselines, indicating its efficiency in reaching the goal location. The reduced navigation times and trajectory lengths can be attributed to the informed decision-making enabled by the multi-modal perception pipeline and the learned covert navigation policy. Our approach achieves the highest exploration coverage percentages, demonstrating its ability to thoroughly explore the environment while navigating towards the goal. The high exploration coverage ensures that the robot can identify potential cover locations and threats more effectively.

Notably, \textit{EnCoMP} significantly reduces the threat exposure percentage compared to the baselines, with values of 10.5\%, 12.0\%, and 14.5\% in Scenarios 1, 2, and 3, respectively. This highlights the effectiveness of our approach in minimizing the robot's visibility to threats and maintaining a high level of covertness during navigation. Furthermore, \textit{EnCoMP} achieves the highest cover utilization percentages, indicating its ability to effectively leverage the available cover objects in the environment. By actively seeking and utilizing cover, our approach enhances the robot's survivability and reduces the risk of detection. 

\begin{figure*}[!htb]
    \centering
    \subfloat[\centering]{{\includegraphics[width=.31\textwidth]{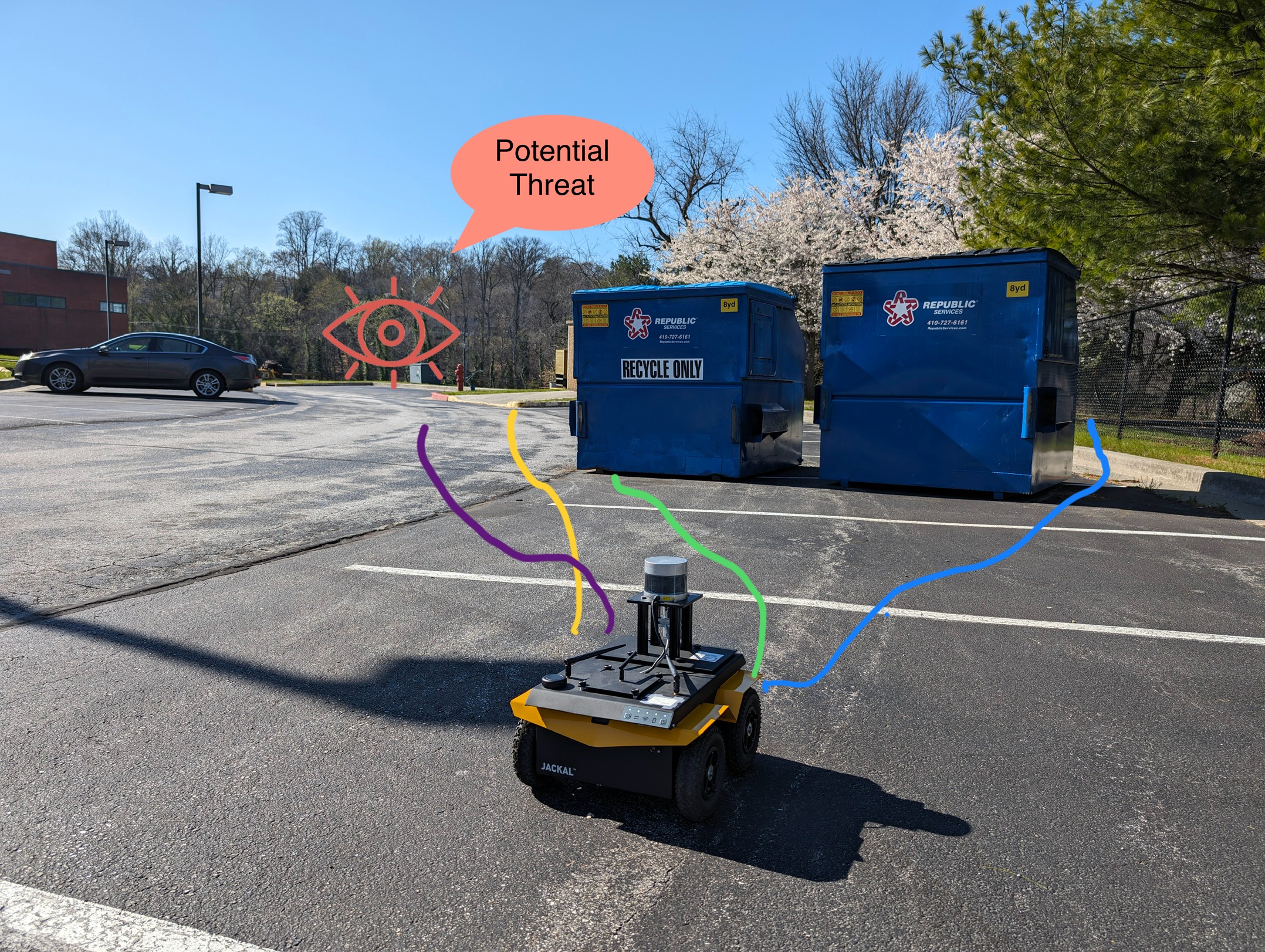}}}%
    \quad % Adjusted for more uniform spacing
    \subfloat[\centering ]{{\includegraphics[width=.31\textwidth]{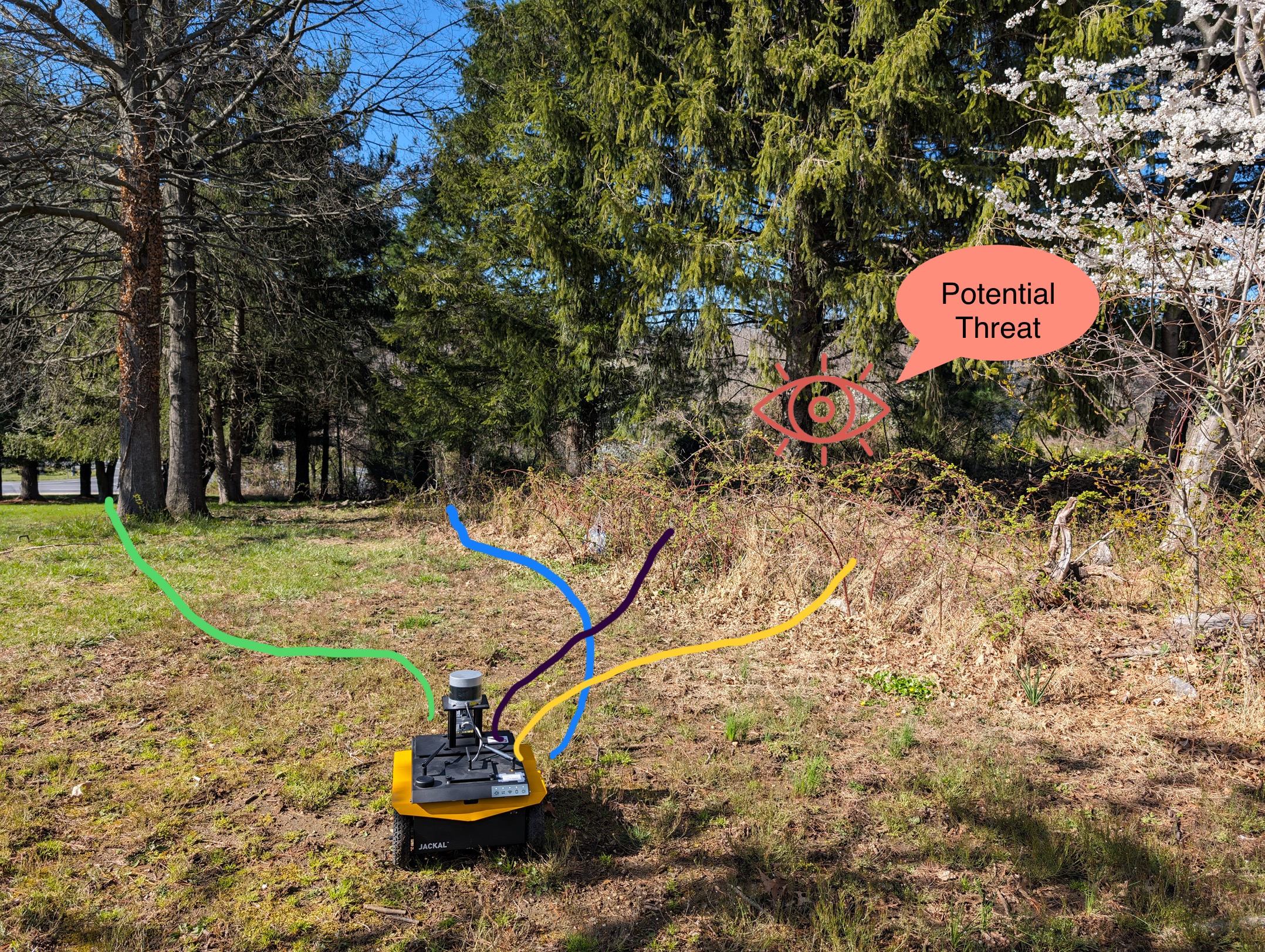} }}%
    \quad % Adjusted for more uniform spacing
    \subfloat[\centering]{{\includegraphics[width=.31\textwidth]{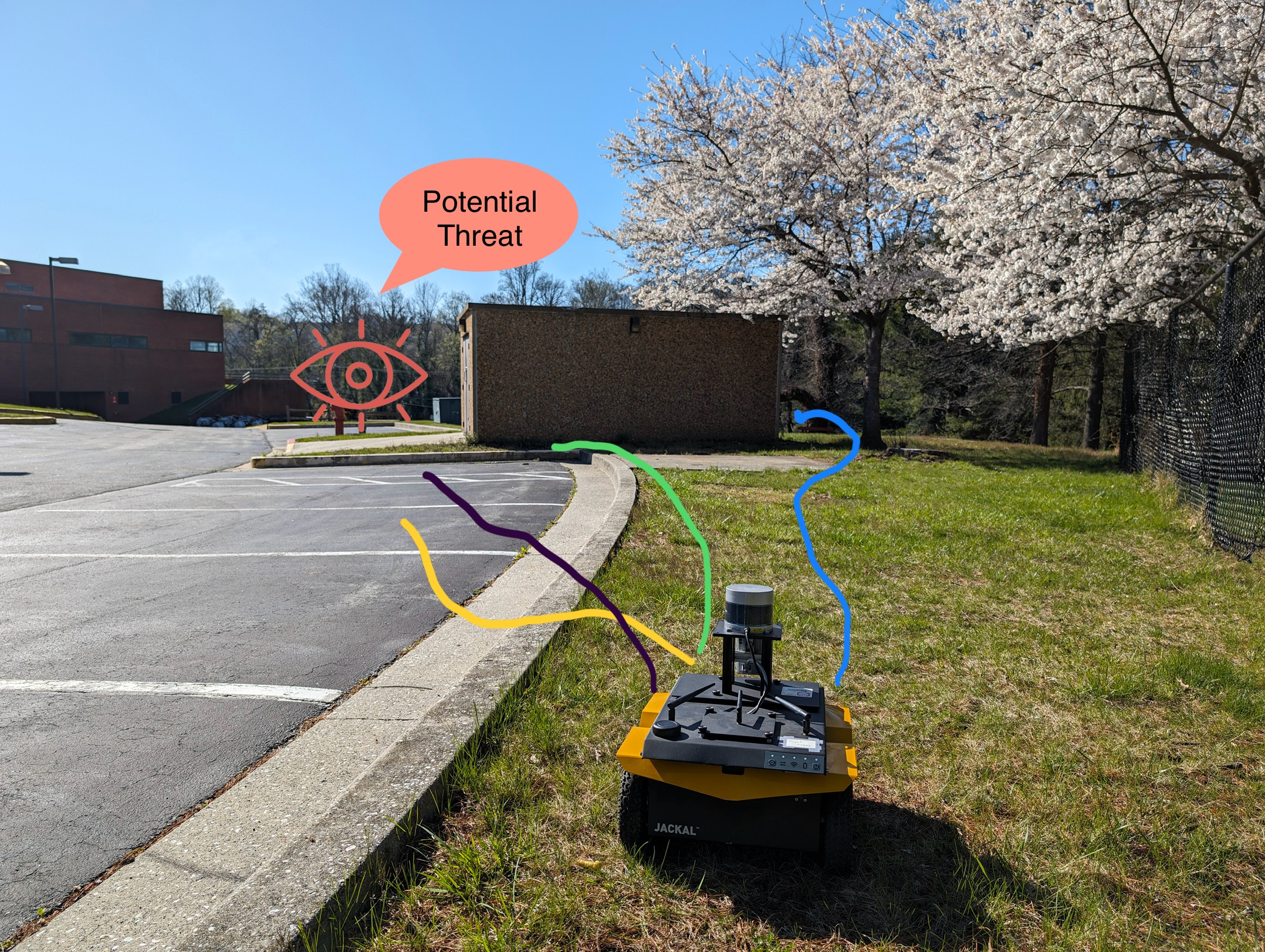}}}%
    
    \caption{Comparison of navigation strategies in diverse outdoor settings (See \ref{sec:testingscenario}). (a) Scenario 1, (b) Scenario 2, (c) Scenario 3. Paths indicate the trajectories taken by different systems, with \textit{EnCoMP} (blue) demonstrating more optimized routes that effectively leverage cover while minimizing threat exposure, in contrast to CoverNav (green), VAPOR (purple), and VERN (yellow).}%
    \label{fig:trajectories}
\end{figure*}

\begin{figure}[!htb]
    \centering
    \includegraphics[width=\linewidth]{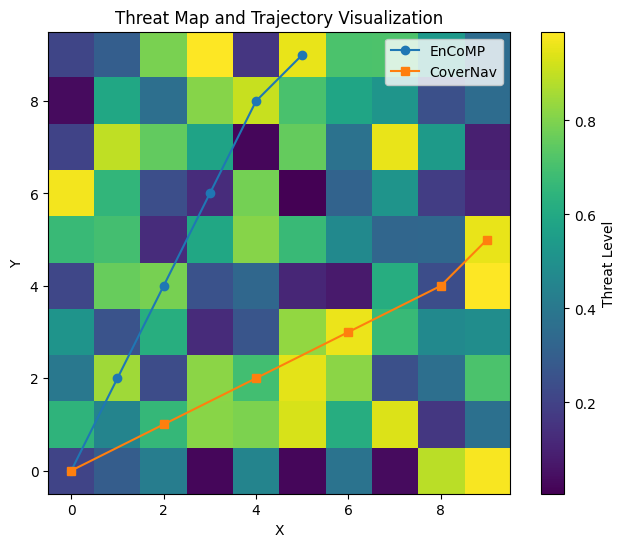}
    \caption{ A visualization of the threat map and the trajectories generated by \textit{EnCoMP} and CoverNav ~\cite{hossain2023covernav} in a representative scenario. The threat map, depicted by the colormap, indicates the level of threat associated with different locations in the environment, with darker colors representing higher threat levels. The \textit{EnCoMP} trajectory (orange line with circular markers) navigates through regions with lower threat levels, demonstrating its ability to identify and prioritize safer routes. In contrast, the CoverNav trajectory (blue line with square markers) traverses through areas with higher threat levels, suggesting a less effective threat avoidance capability. This visualization highlights \textit{EnCoMP}'s superior performance in generating safer paths and minimizing exposure to high-threat regions compared to CoverNav.}
   \label{fig:threat_map_visualization}
\end{figure}

Fig. \ref{fig:trajectories} presents sample navigation trajectories generated by \textit{EnCoMP} and the baseline methods in the three testing scenarios. As evident from the trajectories, \textit{EnCoMP} generates paths that prioritize cover utilization and threat avoidance, while the baseline methods often result in more exposed and less efficient paths. Also, to illustrate the effectiveness of \textit{EnCoMP} in generating safer paths and minimizing exposure to high-threat regions, we present a visualization (Figure~\ref{fig:threat_map_visualization}) of the threat map and the trajectories generated by \textit{EnCoMP} and CoverNav \cite{hossain2023covernav} in a representative scenario.

% \begin{table}[ht]
% \centering
% \scriptsize
% \begin{tabular}{|l|c|c|c|c|}
% \hline
% \textbf{Metrics} & \textbf{Methods} & \textbf{Scen. 1} & \textbf{Scen. 2} & \textbf{Scen. 3} \\
% \hline
% \textbf{Success} & CoverNav & 85 & 83 & 81 \\
% \textbf{Rate (\%)}& VAPOR & 87 & 85 & 84 \\
% & VERN & 88 & 86 & 85 \\
% & \textbf{EnCoMP(Ours)} & \textbf{95} & \textbf{93} & \textbf{91} \\
% \hline
% \textbf{Navigation} & CoverNav & 40.0 & 42.5 & 45.0 \\
% \textbf{Time (s)} & VAPOR & 38.5 & 40.0 & 42.0 \\
% & VERN & 39.0 & 41.0 & 43.5 \\
% & \textbf{EnCoMP(Ours)} & \textbf{32.0} & \textbf{34.5} & \textbf{36.0} \\
% \hline
% \textbf{Trajectory} & CoverNav & 14.0 & 14.5 & 15.0 \\
% \textbf{Length (m)} & VAPOR & 13.5 & 14.0 & 14.5 \\
% & VERN & 13.8 & 14.3 & 14.8 \\
% & \textbf{EnCoMP(Ours)} & \textbf{11.0} & \textbf{12.0} & \textbf{12.5} \\
% \hline
% \textbf{Threat} & CoverNav & 22.5 & 25.0 & 27.5 \\
% \textbf{Exposure (\%)} & VAPOR & 20.0 & 22.5 & 25.0 \\
% & VERN & 18.5 & 21.0 & 23.5 \\
% & \textbf{EnCoMP(Ours)} & \textbf{10.5} & \textbf{12.0} & \textbf{14.5} \\
% \hline
% \textbf{Cover} & CoverNav & 65.0 & 62.5 & 60.0 \\
% \textbf{Utilization (\%)} & VAPOR & 67.5 & 65.0 & 62.5 \\
% & VERN & 70.0 & 67.5 & 65.0 \\
% & \textbf{EnCoMP(Ours)} & \textbf{85.0} & \textbf{82.5} & \textbf{80.0} \\
% \hline
% \end{tabular}
% \caption{Comparative Analysis: \textit{EnCoMP} vs. SOTA methods across three scenarios, highlighting superior performance in metrics such as success rate, navigation time, threat exposure, and cover utilization.}
% \label{tab:real_results}
% \end{table}

\begin{table}[ht]
\centering
\scriptsize
\begin{tabular}{|l|c|c|c|c|}
\hline
\textbf{Metrics} & \textbf{Methods} & \textbf{Scen. 1} & \textbf{Scen. 2} & \textbf{Scen. 3} \\
\hline
\textbf{Success} & CoverNav & 85 & 83 & 81 \\
\textbf{Rate (\%)} & VAPOR & 87 & 85 & 84 \\
& VERN & 88 & 86 & 85 \\
& EnCoMP w/o ATAVE & 90 & 88 & 86 \\
& \textbf{EnCoMP(Ours)} & \textbf{95} & \textbf{93} & \textbf{91} \\
\hline
\textbf{Navigation} & CoverNav & 40.0 & 42.5 & 45.0 \\
\textbf{Time (s)} & VAPOR & 38.5 & 40.0 & 42.0 \\
& VERN & 39.0 & 41.0 & 43.5 \\
& EnCoMP w/o ATAVE & 35.0 & 37.5 & 40.0 \\
& \textbf{EnCoMP(Ours)} & \textbf{32.0} & \textbf{34.5} & \textbf{36.0} \\
\hline
\textbf{Trajectory} & CoverNav & 14.0 & 14.5 & 15.0 \\
\textbf{Length (m)} & VAPOR & 13.5 & 14.0 & 14.5 \\
& VERN & 13.8 & 14.3 & 14.8 \\
& EnCoMP w/o ATAVE & 12.5 & 13.0 & 13.5 \\
& \textbf{EnCoMP(Ours)} & \textbf{11.0} & \textbf{12.0} & \textbf{12.5} \\
\hline
\textbf{Threat} & CoverNav & 22.5 & 25.0 & 27.5 \\
\textbf{Exposure (\%)} & VAPOR & 20.0 & 22.5 & 25.0 \\
& VERN & 18.5 & 21.0 & 23.5 \\
& EnCoMP w/o ATAVE & 15.0 & 17.5 & 20.0 \\
& \textbf{EnCoMP(Ours)} & \textbf{10.5} & \textbf{12.0} & \textbf{14.5} \\
\hline
\textbf{Cover} & CoverNav & 65.0 & 62.5 & 60.0 \\
\textbf{Utilization (\%)} & VAPOR & 67.5 & 65.0 & 62.5 \\
& VERN & 70.0 & 67.5 & 65.0 \\
& EnCoMP w/o ATAVE & 75.0 & 72.5 & 70.0 \\
& \textbf{EnCoMP(Ours)} & \textbf{85.0} & \textbf{82.5} & \textbf{80.0} \\
\hline
\end{tabular}
\caption{Comparative Analysis: \textit{EnCoMP} vs. SOTA methods across three scenarios, highlighting superior performance in metrics such as success rate, navigation time, threat exposure, and cover utilization.}
\label{tab:real_results}
\end{table}

\textbf{Ablation Study for the ATAVE Algorithm:}
We conduct an ablation study to evaluate the impact of the Adaptive Threat-Aware Visibility Estimation (ATAVE) algorithm on the navigation performance of \textit{EnCoMP}. We compare two variants of \textit{EnCoMP}: one with ATAVE and one without ATAVE. The variant without ATAVE relies solely on the cover map and height map for navigation, selecting waypoints based on the highest cover density and lowest elevation. In contrast, \textit{EnCoMP} with ATAVE incorporates dynamic threat assessment and adaptive visibility estimation to generate safer and more efficient trajectories (see Figure \ref{fig:threat_map_visualization}). By considering the potential threats and their visibility, ATAVE enhances the robot's situational awareness and enables it to make informed decisions during covert navigation. The ablation study demonstrates the significant contribution of ATAVE to \textit{EnCoMP}'s overall performance, resulting in higher success rates, lower threat exposure, and improved cover utilization (see Table \ref{tab:real_results}).

\textbf{Computational Efficiency:}
We assess the computational efficiency of \textit{EnCoMP} by comparing its execution times with CoverNav \cite{hossain2023covernav}, a state-of-the-art covert navigation approach. \textit{EnCoMP}'s lightweight architecture and efficient algorithms enable real-time performance, with execution times ranging from 0.05 to 0.08 seconds (12.5-20Hz) on the onboard processing hardware. In contrast, CoverNav requires 0.2 to 0.3 seconds to process the same input data and generate navigation decisions. The computational efficiency of \textit{EnCoMP} is crucial for real-time decision-making and responsive navigation in complex environments. The faster execution times allow the robot to quickly adapt to changing circumstances, such as moving threats or sudden changes in the environment, enhancing its overall performance and survivability during covert missions.
\section{Conclusion, Limitations, and Future Directions}
\label{conclusion}

In this work, we presented \textit{EnCoMP}, an enhanced covert maneuver planning framework that combines LiDAR-based perception and offline reinforcement learning to enable autonomous robots to navigate safely and efficiently in complex environments. Our approach introduces several key contributions, including an advanced perception pipeline that generates high-fidelity cover, threat, and height maps, an offline reinforcement learning algorithm that learns robust navigation policies from real-world datasets, and an effective integration of perception and learning components for informed decision-making. Additionally, we introduced the Adaptive Threat-Aware Visibility Estimation (ATAVE) algorithm, which dynamically assesses and mitigates potential threats in real-time, enhancing the robot's situational awareness and decision-making capabilities.
However, there are a few limitations to our current approach. First, the performance of \textit{EnCoMP} relies on the quality and diversity of the real-world dataset used for training. While we have collected a substantial amount of data, there may still be scenarios or environments that are not well-represented in the dataset. Second, our approach assumes that the environment is static during navigation, which may not always hold in real-world scenarios. Dynamic obstacles or changing terrain conditions could impact the effectiveness of the learned policy. These limitations underscore the need for further research and development to enhance the robustness and adaptability of \textit{EnCoMP} in complex, real-world environments.

Future research directions include extending \textit{EnCoMP} to handle dynamic environments by incorporating online adaptation mechanisms into the reinforcement learning framework. Additionally, we plan to investigate the integration of active perception techniques, such as active mapping or exploration, to allow the robot to actively gather information about the environment during navigation. Another opportunity for future research is to scale up \textit{EnCoMP} to groups or teams of robots, enabling collaborative and coordinated navigation in complex environments. This direction opens up possibilities for multi-robot systems to tackle challenging missions more efficiently and effectively. Furthermore, we plan to validate the performance of \textit{EnCoMP} on a wider range of real-world terrains and scenarios, including more diverse vegetation types, weather conditions, and mission objectives.

\section*{Acknowledgment}
This work has been partially supported by U.S. Army Grant \texttt{\#W911NF2120076} and ONR Grant \texttt{\#N00014-23-1-2119}

\bibliographystyle{unsrt}
\bibliography{bibliography}
\end{document}